\documentclass{article}

\usepackage[preprint]{neurips_2026}

\usepackage[utf8]{inputenc}
\usepackage[T1]{fontenc}
\usepackage{hyperref}
\usepackage{url}
\usepackage{booktabs}
\usepackage{amsfonts}
\usepackage{amsmath} 
\usepackage{amssymb}
\usepackage{nicefrac}
\usepackage{microtype}
\usepackage{xcolor}
\usepackage{graphicx}
\usepackage{multirow}
\usepackage{makecell}
\usepackage{subcaption}
\usepackage{algorithm}
\usepackage{algorithmic}
\usepackage{colortbl}
\usepackage{array}

\usepackage{enumitem}
\usepackage{amsthm}
\newtheorem{proposition}{Proposition}
\newtheorem{remark}{Remark}

\providecommand{\answerYes}[1][]{\textcolor{blue}{Yes}}
\providecommand{\answerNo}[1][]{\textcolor{orange}{No}}
\providecommand{\answerNA}[1][]{\textcolor{gray}{N/A}}
\newcommand{\graftgnn}{\textsc{Graft}}
\newcommand{\gnnx}{\textsc{GnnXemplar}}
\newcommand{\fidplus}{\ensuremath{\text{Fid}^+}}
\newcommand{\fidminus}{\ensuremath{\text{Fid}^-}}

\newcommand{\xmark}{\ensuremath{\times}}
\DeclareMathOperator*{\argmin}{arg\,min}
\DeclareMathOperator*{\argmax}{arg\,max}
\definecolor{rowgray}{gray}{0.93}

\title{\graftgnn{}: Auditing Graph Neural Networks via Global Feature Attribution}

\author{%
  Rishi Raj Sahoo$^{1,2}$ \quad Subhankar Mishra$^{1,2}$ \\
  $^{1}$National Institute of Science Education and Research (NISER), Bhubaneswar, India \\
  $^{2}$Homi Bhabha National Institute, Mumbai, India \\
  \texttt{\{rishiraj.sahoo, smishra\}@niser.ac.in}
}

\begin{document}
\maketitle

\begin{abstract}
Graph Neural Networks (GNNs) achieve strong performance on node classification tasks but remain difficult to interpret, particularly with respect to \emph{which input features} drive their predictions.  Existing global GNN explainers operate at the \emph{structural} level identifying recurring subgraph motifs, but none explain model behaviour \emph{globally} at the level of input node attributes. We propose \graftgnn{}, a post-hoc global explanation framework that identifies class-level feature importance profiles for GNNs. The method combines diversity-guided exemplar selection, Integrated Gradients-based attribution, and aggregation to construct a global view of feature influence for each class, which can be further expressed as concise natural language rules using a large language model with self-refinement. We evaluate \graftgnn{} across multiple datasets, architectures, and experimental settings, demonstrating its effectiveness in capturing model-relevant features, supporting bias analysis, and enabling feature-efficient transfer learning. In addition, we introduce a structured human evaluation protocol to assess the interpretability of generated rules along dimensions such as accuracy and usefulness. Our results suggest that \graftgnn{} provides a practical and interpretable approach for analysing feature-level behaviour in GNNs, bridging quantitative attribution with human-understandable explanations.
\end{abstract}

\section{Introduction}
\label{sec:introduction}

Graph Neural Networks achieve state-of-the-art accuracy on node classification
across citation networks, co-authorship graphs, product graphs, and social
networks~\citep{kipf2017semi,velickovic2018graph,hamilton2017inductive,xu2019powerful}.
Yet their predictions remain largely opaque.
A practitioner cannot easily answer: \emph{which node attributes drive the model's
class decisions?}
This opacity is not merely inconvenient: a GNN trained for citation classification
may exploit irrelevant author-name tokens, or a product-recommendation model may
rely on price artefacts injected at training time.
Without a way to audit \emph{which features a GNN actually uses for each class},
such shortcuts go undetected until deployment.

\paragraph{Model auditing gap.}
We propose \emph{feature shortcut detection} as the primary evaluation target for
global GNN explainers.
Given a trained GNN, does the explainer identify features the model relies on,
rather than features merely correlated with class labels in the training set?
We formalise this through controlled bias-injection experiments: a synthetic
binary feature is correlated with a target class and injected into the graph;
a trustworthy explainer must surface this feature prominently.
No existing global GNN explainer has demonstrated this capability.

\begin{figure*}[t]
\centering
\includegraphics[width=\linewidth]{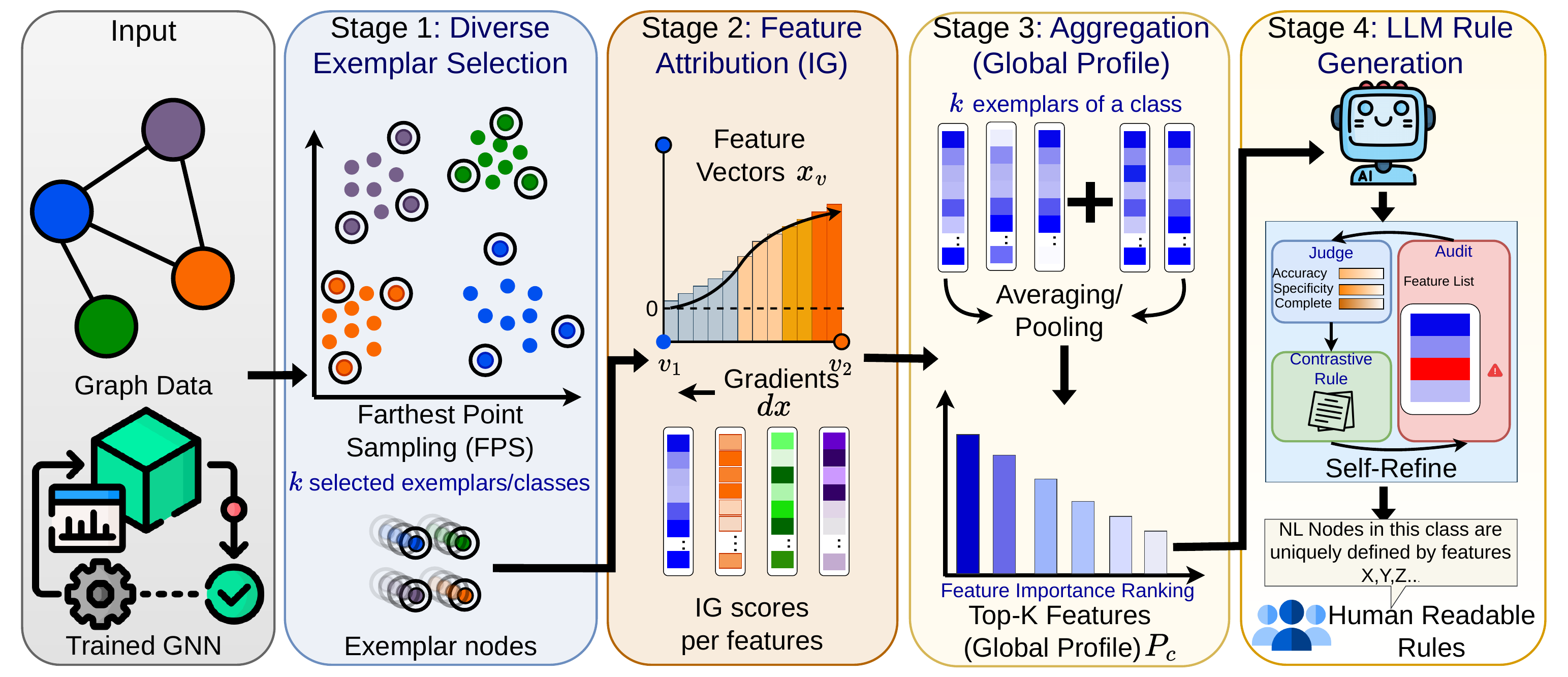}
\caption{\graftgnn{} pipeline. \textbf{Input}: trained GNN $f$ and graph $(G,X)$.
\textbf{Stage~1} selects $k$ diverse exemplar nodes per class via Farthest Point Sampling in
embedding space.
\textbf{Stage~2} computes Integrated Gradients for each exemplar: a zero baseline,
50 Riemann interpolation steps, yielding $\mathrm{IG}_e \in \mathbb{R}^d$.
\textbf{Stage~3} aggregates $|\mathrm{IG}|$ over exemplars and ranks features by mean importance,
producing class profile $\mathcal{P}_c$.
\textbf{Stage~4} verbalises $\mathcal{P}_c$ as a natural language rule $r_c$
via an LLM with self-refinement, which is human-readable.}
\label{fig:pipeline}
\end{figure*}

\paragraph{This work.}
We fill the global feature-level explainability gap with \graftgnn{}
(\textbf{G}lobal g\textbf{R}aph \textbf{A}ttribute \textbf{F}eature a\textbf{T}tribution),
a post-hoc, model-agnostic method that works with any GNN producing node embeddings.
The pipeline has four stages as illustrated in Figure \ref{fig:pipeline}:
(1) select $k$ maximally diverse class-representative nodes via Farthest Point Sampling;
(2) compute Integrated Gradients~\citep{sundararajan2017axiomatic} per exemplar;
(3) aggregate attributions into a ranked class-level feature importance profile; and
(4) verbalise the profile as natural language rules via an LLM.
GNN explainability methods split along locality (local vs.\ global) and granularity
(structural vs.\ feature-level) axes.
Global structural methods now exist: \gnnx{}~\citep{azzeh2025gnnxemplar} identifies
representative subgraph patterns per class; CIFlow-GNN~\citep{ciflow2026} uses
causal flows over graph topology.
The \emph{global feature-level} quadrant is unoccupied; \graftgnn{} fills it.

Beyond profiles, we introduce three analysis tools:
(i) \emph{Cross-architecture consensus}: what fraction of top-$K$ features are
identified independently by $\geq 3$ of 4 architectures?
(ii) \emph{Contrastive profiles}: for class~$c$, which features have high importance
for~$c$ but not for any other class?
(iii) \emph{Jaccard stability}: how reproducible is the top-$K$ feature set across seeds?

\paragraph{Contributions.}
\begin{enumerate}
  \item \textbf{Auditing application}: the first demonstration of feature shortcut
        detection for GNNs via global attribution: injected spurious features are
        recovered within top-3 across all 11 (dataset, architecture) pairs tested
        (GCN and GAT architectures; \S\ref{sec:bias}), rank~1 in 9 of 11.
  \item \textbf{Method}: \graftgnn{}, the first global, feature-level, post-hoc GNN explainer
        with natural language rule generation (\S\ref{sec:method}).
  \item \textbf{Discriminative evaluation}: \graftgnn{}-selected features (1\%--17\% of
        the feature space on the evaluated benchmarks) achieve competitive transfer performance, outperforming frequency-based selection on several datasets while remaining comparable on others (\S\ref{sec:transfer}).
  \item \textbf{Systematic evaluation}: 260 runs across 13 datasets, 4 architectures,
        5 seeds; three novel analysis tools quantify explanation quality beyond fidelity.
\end{enumerate}

\section{Related Work}
\label{sec:related}

\paragraph{Local GNN explainability.}
GNNExplainer~\citep{ying2019gnnexplainer} learns per-instance edge/feature masks;
PGExplainer~\citep{luo2020parameterized} parametrises masks for generalisation.
Both target individual predictions; their outputs cannot be straightforwardly
aggregated to class-level feature attributions.
Per-node Integrated Gradients and SHAP~\citep{lundberg2017unified} produce
scalar feature attributions for each node, but a class-level summary requires
choosing which nodes to include and how to pool, choices that are ad-hoc
and that strongly affect the result.
\emph{From Nodes to Narratives}~\citep{fromnodestonarratives2025} generates
LLM-based natural language explanations at the per-node (local) level.

\paragraph{Global GNN explainability.}
\gnnx{}~\citep{azzeh2025gnnxemplar} is the closest prior work: it selects
structural exemplars and uses an LLM to describe recurring subgraph patterns.
Its attribution target is graph structure; no feature importance profile is produced.
CIFlow-GNN~\citep{ciflow2026} identifies globally influential structural components
via causal intervention flows.
Global Shapley methods~\citep{shapley_gnn2025} attribute importance to activation
patterns in intermediate layers rather than raw input features.
\graftgnn{} is complementary to \gnnx{}: the two methods can be combined
("\graftgnn{}+", see \S\ref{sec:discussion}) to produce explanations covering
both structural and feature-level aspects of a class.
Table~\ref{tab:related} summarises the landscape.

\begin{table}[h]
\centering
\caption{Comparison of global GNN explanation methods.
\graftgnn{} uniquely combines global scope, input-feature attribution, and NL rules.}
\label{tab:related}
\small
\begin{tabular}{lccc}
\toprule
Method & Global? & Feature-level? & NL rules? \\
\midrule
GNNExplainer~\citep{ying2019gnnexplainer}                 & \xmark     & \checkmark (local) & \xmark     \\
\gnnx{}~\citep{azzeh2025gnnxemplar}                       & \checkmark & \xmark (structural) & \checkmark \\
CIFlow-GNN~\citep{ciflow2026}                              & \checkmark & \xmark (structural) & \xmark     \\
From Nodes to Narratives~\citep{fromnodestonarratives2025} & \xmark     & \checkmark (local)  & \checkmark \\
Shapley for GNNs~\citep{shapley_gnn2025}                  & \checkmark & Partial (activations) & \xmark   \\
\midrule
\textbf{\graftgnn{} (ours)}                                   & \checkmark & \checkmark          & \checkmark \\
\bottomrule
\end{tabular}
\end{table}

\paragraph{Feature attribution.}
Integrated Gradients~\citep{sundararajan2017axiomatic} satisfies completeness
(attributions sum to $f(x) - f(\mathbf{0})$) and sensitivity (features affecting
output receive non-zero attribution).
\graftgnn{} uses IG as a subroutine; any scalar-per-feature attribution method
(Grad$\times$Input~\citep{shrikumar2017learning}, SHAP~\citep{lundberg2017unified})
can be substituted.

\paragraph{Exemplar-based explanations.}
ProtGNN~\citep{zhang2022protgnn} learns prototypical subgraphs embedded in
the model architecture (intrinsic, not post-hoc).
\graftgnn{} is post-hoc, requiring no architectural changes.
FPS-based diversity selection from~\gnnx{} is adapted here from the structural
to the feature attribution domain.

\paragraph{Shortcut and bias detection.}
Spurious correlations in neural networks have been studied
in vision~\citep{geirhos2020shortcut} and NLP~\citep{nam2020learning}.
To our knowledge, \graftgnn{} is the first method to demonstrate feature shortcut
detection for GNNs via global feature attribution.

\section{Method}
\label{sec:method}

\subsection{Problem Formulation}

Let $G = (\mathcal{V}, \mathcal{E}, X)$ be a graph with node feature matrix
$X \in \mathbb{R}^{|\mathcal{V}| \times d}$ and labels $y \in \{1,\ldots,C\}^{|\mathcal{V}|}$.
Let $f$ be a trained GNN node classifier with $f_c(x, G)$ denoting the
class-$c$ logit for a node with features $x$.

\textbf{Goal.} For each class $c$, produce a ranked profile
$\mathcal{P}_c = [(i_1, s_1), \ldots, (i_K, s_K)]$
of the $K$ input features most responsible for $f$'s classification of nodes
as class~$c$, where $s_j$ is feature~$i_j$'s importance score.
Produce a natural language rule $r_c$ describing the class in terms of these features.

\subsection{\graftgnn{}}

\paragraph{Stage 1: Diversity-guided exemplar selection.}
Let $h_v \in \mathbb{R}^{d_h}$ be the penultimate-layer embedding of node~$v$.
Rather than averaging attributions over all class nodes (expensive and dominated
by the centroid, which reduces explanation diversity), we select $k$ maximally
\emph{diverse} exemplars via Farthest Point Sampling (FPS)~\citep{eldar1997farthest}:
\begin{equation}
  e_1 = \argmin_{v \in \mathcal{V}_c} \|h_v - \bar{h}_c\|_2, \qquad
  e_{j+1} = \argmax_{v \in \mathcal{V}_c} \min_{i \leq j} \|h_v - h_{e_i}\|_2,
\end{equation}
where $\bar{h}_c$ is the class centroid.
FPS guarantees maximum minimum inter-point distance, covering the class manifold
without redundancy.
Crucially, because FPS is a \emph{deterministic} function of fixed trained embeddings,
profiles built from FPS exemplars are highly reproducible across random seeds
(cf.\ Jaccard stability, \S\ref{sec:stability}), whereas profiles built from
randomly-sampled subsets of class nodes exhibit substantially higher variance.
We use $k=10$ exemplars per class throughout.
Separately, $K=20$ top features are selected for fidelity evaluation; ablations
over both $k$ and $K$ are in Appendix~\ref{app:ablation}.

\paragraph{Stage 2: Integrated Gradients.}
Integrated Gradients provides a path-integrated measure of feature influence, capturing non-linear interactions between input dimensions. In the context of GNNs, this is particularly important as feature contributions are mediated through message passing and aggregation. The IG formulation ensures that features contributing through higher-order neighborhood effects are still reflected in the attribution signal.
For exemplar $e \in \mathcal{E}_c$, with zero baseline (absence of features):
\begin{equation}
  \mathrm{IG}_e[i] = x_e[i] \cdot \int_0^1
    \frac{\partial f_c(\alpha \cdot x_e, G)}{\partial x[i]} \,d\alpha,
  \label{eq:ig}
\end{equation}
approximated with 50 Gauss-Legendre steps using Captum~\citep{kokhlikyan2020captum}.
The zero baseline is natural for bag-of-words and TF-IDF features where $0$
means word absent.\footnote{Runtime: on Cora/GCN (2,708 nodes, 70 exemplars,
50 IG steps) the attribution stage completes in $\approx$28s on a single CPU core.
A frequency-based profile takes $<$1s; the IG overhead is justified by
stability, consensus, and bias-detection capabilities unavailable from frequency.}

\paragraph{Stage 3: Class-level aggregation.}
The aggregation step transforms local, instance-level explanations into a global class-level representation. This step is critical: naive averaging over all nodes would dilute informative signals due to redundancy and class imbalance. By combining FPS-based diversity with aggregation, GRAFT ensures that the resulting profile captures the \emph{support of the class manifold} rather than its centroid alone. The default aggregation is a simple mean:
\begin{equation}
  \mu_c = \frac{1}{|\mathcal{E}_c|} \sum_{e \in \mathcal{E}_c} |\mathrm{IG}_e|
  \in \mathbb{R}^d.
  \label{eq:agg}
\end{equation}
We also introduce a \emph{confidence-weighted} variant that upweights exemplars
for which the GNN is more certain:
\begin{equation}
  \mu_c^{\mathrm{cw}} = \frac{\sum_{e \in \mathcal{E}_c} p_c(e)\,|\mathrm{IG}_e|}
                             {\sum_{e \in \mathcal{E}_c} p_c(e)},
  \label{eq:agg_cw}
\end{equation}
where $p_c(e) = \mathrm{softmax}(f(x_e,G))[c]$ is the GNN's predicted probability
for class~$c$ on exemplar~$e$.
High-confidence exemplars are the most unambiguous class representatives;
their attributions are more reliable signals.
The top-$K$ indices by $\mu_c$ (or $\mu_c^{\mathrm{cw}}$) form the core profile $\mathcal{P}_c$.

\begin{remark}[Aggregate completeness]
IG satisfies per-exemplar completeness: $\sum_i \mathrm{IG}_e[i] = f_c(x_e, G) - f_c(\mathbf{0}, G)$.
Taking the mean over $\mathcal{E}_c$ gives
$\langle \mu_c, \mathbf{1} \rangle = \frac{1}{|\mathcal{E}_c|}
\sum_{e} [f_c(x_e, G) - f_c(\mathbf{0}, G)]$,
the average \emph{gap} between exemplar predictions and the all-zero baseline.
This aggregate quantity is a meaningful measure of how strongly the class is
"supported" by its top features, grounding $\mu_c$ in the model's actual output differences.
\end{remark}

\begin{proposition}[FPS profile approximation]\label{prop:fps}
Let $|\mathrm{IG}|:\mathcal{V}_c\!\to\!\mathbb{R}^d$ be $L$-Lipschitz with respect to the GNN
embedding: $\big\||\mathrm{IG}_u|-|\mathrm{IG}_v|\big\|_\infty \leq L\|h_u - h_v\|_2$
for all $u,v\in\mathcal{V}_c$.
Let $r_k$ be the FPS coverage radius and assume balanced Voronoi cells.
Then
\begin{equation}
  \bigl\|\mu_c^{} - \mu_c^{*}\bigr\|_\infty \;\leq\; L\cdot r_k,
  \label{eq:fps_bound}
\end{equation}
where $\mu_c^{*} = \frac{1}{|\mathcal{V}_c|}\sum_{v\in\mathcal{V}_c}|\mathrm{IG}_v|$
is the oracle profile over all class nodes.
Since FPS is a 2-approximation to the optimal $k$-centre
radius~\citep{gonzalez1985clustering}, $r_k \leq 2r_k^*$, and increasing $k$
monotonically tightens the bound.
\end{proposition}

\begin{proposition}[Fidelity-attribution correspondence]\label{prop:fid}
Assume a \emph{linear decoder}: $f_c(x,G) = a_c^\top x + b_c$ where
$a_c = B^\top w_c$ for a fixed graph-dependent matrix $B$ and bias $b_c$
(valid for any GNN with a linear readout head, including GCNConv which has
a bias by default).
Then for node $v$ with top-$K$ profile $T_c$:
\begin{align}
  f_c(x_v,G) - f_c(x_v^{-T_c},G) &= \sum_{i \in T_c}\mathrm{IG}_v[i],
  \label{eq:fid_plus}\\
  f_c(x_v^{T_c},G) - f_c(\mathbf{0},G) &= \sum_{i \in T_c}\mathrm{IG}_v[i],
  \label{eq:fid_minus}
\end{align}
where $x_v^{T_c}$ zeros all features outside $T_c$.
Both quantities equal $\sum_{i\in T_c}\mathrm{IG}_v[i]$: the profile
captures exactly that portion of the class-$c$ logit.
Furthermore, under the \emph{class-positive} assumption ($\mathrm{IG}_v[i]\geq 0$
for all $i\in T_c$, which holds when top-$K$ features positively support class~$c$,
e.g.\ for binary bag-of-words with a well-trained classifier),
$\sum_{i\in T_c(K)}\mathrm{IG}_v[i]$ is non-decreasing in $K$.
\end{proposition}

\noindent Proofs are in Appendix~\ref{app:proofs}.
Proposition~\ref{prop:fps} directly justifies why more exemplars ($\uparrow k$)
improve profile quality; Proposition~\ref{prop:fid} provides a logit-level
interpretation of fidelity as attribution capture, and explains the monotone
$\mathrm{Fid}^-$ vs $K$ trend in our ablation (Appendix~\ref{app:ablation})
under the mild class-positive assumption satisfied by bag-of-words features.

Contrastive profiles (subtracting the max attribution of other classes from $\mu_c[i]$)
are defined in Appendix~\ref{app:contrastive} and used for qualitative analysis only.

\paragraph{Stage 4: Natural language rules.}
The top-15 features from $\mathcal{P}_c$, with importance scores, are passed to
an LLM (Claude~Sonnet~4.6 / Gemini / Ollama) with dataset-specific context,
using temperature~$=0.2$.
The same model acts as both generator and self-refinement judge: an initial rule
is generated from the feature list, then a single self-refinement pass re-examines
the rule against the features and rewrites it if any aspect (accuracy, specificity,
completeness) can be improved, returning it unchanged otherwise (exact prompts in
Appendix~\ref{app:rules}).
All quantitative evaluations use $\mathcal{P}_c$ directly; Stage~4 adds a
human-readable layer for practitioners.

\subsection{Analysis Tools}

\paragraph{Cross-architecture consensus.}
Let $T_c^{(m)}$ be the top-$K$ feature indices from model~$m$.
\begin{equation}
  \mathrm{Consensus}_c(\tau) =
  \frac{\bigl|\{i : |\{m : i \in T_c^{(m)}\}| \geq \tau\}\bigr|}{K}.
\end{equation}
We use $\tau = 3$ (out of 4 architectures).
High consensus means the explanation is architecture-invariant.

\paragraph{Jaccard stability.}
Let $T_c^{(s)}$ be the top-$K$ set from seed~$s$.
\begin{equation}
  J_c = \frac{1}{\binom{N_s}{2}}
  \sum_{s < s'} \frac{|T_c^{(s)} \cap T_c^{(s')}|}{|T_c^{(s)} \cup T_c^{(s')}|}.
\end{equation}
$J_c = 1$ means perfectly reproducible explanations; $J_c = 0$ means total instability.

\section{Experiments}
\label{sec:experiments}

\subsection{Setup}

\paragraph{Datasets.}
13 node-classification benchmarks (Table~\ref{tab:datasets}, Appendix~\ref{app:datasets}):
\textbf{citation networks} (Cora, CiteSeer, PubMed~\citep{mccallum2000automating,namata2012query}),
\textbf{co-authorship} (Coauthor-CS, Coauthor-Physics~\citep{shchur2018pitfalls}),
\textbf{Amazon products} (Computers, Photo~\citep{shchur2018pitfalls}),
\textbf{Wikipedia heterophilic} (Chameleon, Squirrel~\citep{rozemberczki2021multi}),
\textbf{WebKB} (Wisconsin, Cornell, Texas~\citep{pei2020geom}),
and \textbf{Actor}~\citep{pei2020geom}.

\paragraph{Models, reproducibility, and compute.}
We evaluate four standard GNN architectures: GCN~\citep{kipf2017semi}, GAT~\citep{velickovic2018graph}, GraphSAGE~\citep{hamilton2017inductive}, and GIN~\citep{xu2019powerful}, each configured with two layers, 64 hidden dimensions, and trained for 500 epochs using the Adam optimiser. For reproducibility, we use five random seeds (0--4) controlling weight initialisation and data splits, and report all results as mean~$\pm$~standard deviation across seeds, yielding a total of $13 \times 4 \times 5 = 260$ runs. Experiments are conducted on a server with 4$\times$ NVIDIA RTX~6000 Ada Generation GPUs (48\,GB VRAM each) and 128 CPU cores; each run takes under 3 minutes on a single GPU, and the full evaluation completes in under 2 hours using parallel execution.

\paragraph{Metrics.}
\textbf{Primary}: (i)~\emph{Bias detection rank}: position of an injected spurious
feature in the target class profile; lower is better.
(ii)~\emph{Transfer classifier accuracy} (GRAFT-LR vs Freq-LR vs Full-LR): logistic
regression trained on the identified feature subset.
\textbf{Secondary}: \fidminus{} (accuracy retaining top-$K$ features) and
\fidplus{} (accuracy drop when top-$K$ features are masked), $K=20$ throughout.
\textbf{Structural}: Jaccard stability (explanation reproducibility across seeds)
and cross-architecture consensus (fraction of architecture-invariant features).

\paragraph{Baselines.}
\textbf{Frequency}: top-$K$ features by per-class mean value over training nodes.
\textbf{Random}: $K$ uniformly random features.
Neither baseline uses the trained model; frequency requires only the training set. 

\subsection{Bias Detection: Model Auditing}
\label{sec:bias}

We evaluate whether \graftgnn{} can serve as a \emph{model auditing} tool by
testing its ability to detect injected feature shortcuts, a controlled proxy
for real-world spurious correlations.
A trustworthy global explainer must surface features the GNN actively relies on,
not merely features frequent in the training data; the two coincide only when
the model has learned the same correlations as the data distribution.
This experiment disambiguates the two.

\paragraph{Protocol.}
For dataset $\mathcal{D}$, we append a synthetic binary feature $z$ with
$P(z_v=1 \mid y_v = c_0) = 1-\sigma$ and
$P(z_v=1 \mid y_v \neq c_0) = \sigma$,
retrain a fresh GNN for 300 epochs, then run \graftgnn{}.
\emph{Detection} succeeds if the injected feature appears in top-20 for class~$c_0$.

\begin{table}[t]
\centering
\caption{Bias detection results ($\sigma=0.05$, seed~42). Rank = position of injected feature (1 = top). "Other" = number of non-target classes where the feature appears.}
\label{tab:bias}
\scriptsize
\setlength{\tabcolsep}{3pt}
\begin{tabular}{llccc|llccc|llccc}
\toprule
\multicolumn{5}{c}{Citation} & \multicolumn{5}{c}{Heterophilic} & \multicolumn{5}{c}{WebKB} \\
\cmidrule(lr){1-5} \cmidrule(lr){6-10} \cmidrule(lr){11-15}
Data & Model & Detected? & Rank & Other & Data & Model & Detected? & Rank & Other & Data & Model & Detected? & Rank & Other \\
\midrule
Cora     & GCN & \checkmark & 1 & 4/6 & Actor    & GCN & \checkmark & 2 & 4/4 & Wisconsin & GCN & \checkmark & 1 & 1/4 \\
CiteSeer & GCN & \checkmark & 1 & 2/5 & Actor    & GAT & \checkmark & 1 & 4/4 & Wisconsin & GAT & \checkmark & 1 & 1/4 \\
CiteSeer & GAT & \checkmark & 1 & 4/5 & Squirrel & GCN & \checkmark & 3 & 4/4 &          &     &            &   &     \\
PubMed   & GCN & \checkmark & 1 & 1/2 & Squirrel & GAT & \checkmark & 1 & 2/4 &          &     &            &   &     \\
PubMed   & GAT & \checkmark & 1 & 2/2 &          &     &            &   &     &          &     &            &   &     \\
\bottomrule
\end{tabular}
\end{table}

Across all 11 tested (dataset, architecture) pairs (Table~\ref{tab:bias}),
\graftgnn{} detects the injected feature within the top-3 of the target class profile in every case.
In 9 of 11 pairs, the injected feature appears at rank~1.
Actor (GCN) places it at rank~2, and Squirrel (GCN) at rank~3; both are topology-dominated
heterophilic datasets where feature signals are weaker overall.
In some cases, the injected feature also surfaces in non-target class profiles
(e.g., all 4 non-target classes on Actor and Squirrel/GCN), because a low-noise
correlation ($\sigma=0.05$) induces some model reliance on the spurious feature beyond
the target class; the target class consistently shows a stronger signal.

Noise robustness ($\sigma \in \{0.05$--$0.40\}$) and multi-class results
are in Appendix~\ref{app:bias_extra}; detection holds in all tested settings.

\graftgnn{}'s ability to surface injected shortcuts suggests a practical audit workflow:
run \graftgnn{}, inspect the top features per class, and flag any feature with
high importance but no clear semantic relationship to the class label.
To our knowledge, this is the first demonstration of feature shortcut detection for GNNs
via global attribution.

\subsection{Transfer Classifier Evaluation}
\label{sec:transfer}

The cleanest test of whether \graftgnn{} identifies genuinely discriminative
features is a \emph{transfer classifier}: train a logistic regression (LR)
on the \graftgnn{}-selected feature subset and measure its accuracy against
the same model trained on frequency-baseline features or on all features.
If gradient attribution adds value beyond raw class statistics, GRAFT-LR should
outperform Freq-LR even when both use the same number of features.
Table~\ref{tab:transfer} reports results (mean~$\pm$~std over 5 seeds).

\begin{table}[t]
\centering
\caption{Transfer classifier accuracy (mean~$\pm$~std, 5 seeds, best architecture per dataset).
\graftgnn-LR: LR on \graftgnn{}-selected features (top-$K$ per class, unioned).
Freq-LR: frequency-baseline features.
Full-LR: all features.
Compress: mean fraction of features used.
\textbf{Bold}: best LR method.
Full results in Appendix~\ref{app:full_transfer}.}
\label{tab:transfer}
\small
\setlength{\tabcolsep}{3.5pt}
\begin{tabular}{llcccc>{\columncolor{rowgray}}c}
\toprule
Dataset & Arch & \graftgnn-LR & Freq-LR & Full-LR & GNN & Compress \\
\midrule
Cora             & GAT  & $\mathbf{0.496}\pm0.028$ & 0.486 & 0.468 & 0.785 & 8\% \\
Coauthor-Physics & GCN  & $0.708\pm0.027$ & 0.784 & \textbf{0.881} & 0.935 & \textbf{1\%} \\
Photo            & SAGE & $0.686\pm0.033$ & 0.702 & \textbf{0.722} & 0.895 & 17\% \\
Squirrel         & SAGE & $0.271\pm0.009$ & 0.185 & \textbf{0.281} & 0.315 & 5\% \\
Cornell          & SAGE & $\mathbf{0.724}\pm0.027$ & 0.703 & 0.703 & 0.751 & 4\% \\
Actor            & SAGE & $\mathbf{0.356}\pm0.004$ & 0.342 & 0.334 & 0.323 & 9\% \\
\bottomrule
\end{tabular}
\end{table}

\graftgnn-LR exceeds Freq-LR on four of the six datasets (Cora, Squirrel, Cornell, Actor),
including a $+0.09$ margin on Squirrel ($0.271$ vs $0.185$) where topology-dominated GNNs extract latent feature structure that class-frequency statistics miss.
On Cora, Cornell, and Actor, \graftgnn-LR also exceeds Full-LR while using
4--9\% of features; on Squirrel, it nearly matches Full-LR ($0.271$ vs $0.281$)
at 5\% compression.
On Coauthor-Physics (8,415 features), \graftgnn{} achieves $70.8\%$ accuracy at
$<\!1\%$ compression ($80\%$ of Full-LR); on Photo (745 features), it is within
$0.016$ of Freq-LR at 19\% compression.
These two cases reveal a limit of gradient attribution on dense feature spaces:
when thousands of features all contribute weakly, rank-20 selection loses discriminative
signal that frequency statistics can be cheaply recovered.

\subsection{Fidelity Results}

\fidminus{} measures the fraction of GNN accuracy retained when using \emph{only}
the \graftgnn{}-identified features, providing a second lens on explanation quality
complementary to the transfer and auditing experiments above.
Table~\ref{tab:fidelity_main} reports \fidminus{} for all 13 datasets across all four
architectures (mean\,$\pm$\,std over 5 seeds).
\fidplus{} results (accuracy drop when the top-$K$ features are masked) are in
Appendix~\ref{app:full_fidelity}.

\paragraph{Aggregation variant.}
In addition to mean aggregation, we evaluate a confidence-weighted variant that assigns higher weight to nodes with more stable attribution scores. This variant consistently improves fidelity across datasets (Appendix \ref{app:ablation_csfps}). We retain mean aggregation as the default for simplicity and interpretability.

\begin{table*}[t]
\centering
\caption{\fidminus{} (mean\,$\pm$\,std, 5 seeds): fraction of GNN accuracy retained using only
the top-$K{=}20$ GRAFT-identified features. \textbf{Bold}: best architecture per dataset.}
\label{tab:fidelity_main}
\small
\setlength{\tabcolsep}{5pt}
\begin{tabular}{l cccc}
\toprule
Dataset & GCN & GAT & SAGE & GIN \\
\midrule
\rowcolor{rowgray}
\multicolumn{5}{l}{\textit{Citation networks}} \\
Cora             & $\mathbf{0.793}\pm0.009$ & $0.699\pm0.121$ & $0.776\pm0.099$ & $0.647\pm0.074$ \\
CiteSeer         & $0.674\pm0.100$           & $0.733\pm0.049$ & $\mathbf{0.775}\pm0.034$ & $0.439\pm0.051$ \\
PubMed           & $0.695\pm0.125$           & $0.557\pm0.183$ & $\mathbf{0.806}\pm0.116$ & $0.639\pm0.147$ \\
\midrule
\rowcolor{rowgray}
\multicolumn{5}{l}{\textit{Co-authorship graphs}} \\
Coauthor-CS      & $0.792\pm0.100$ & $0.671\pm0.129$ & $\mathbf{0.903}\pm0.053$ & $0.644\pm0.055$ \\
Coauthor-Physics & $0.785\pm0.107$ & $0.573\pm0.198$ & $\mathbf{0.815}\pm0.066$ & $0.498\pm0.062$ \\
\midrule
\rowcolor{rowgray}
\multicolumn{5}{l}{\textit{Amazon product graphs}} \\
Computers        & $0.432\pm0.269$ & $0.140\pm0.046$ & $\mathbf{0.541}\pm0.218$ & $0.106\pm0.008$ \\
Photo            & $\mathbf{0.466}\pm0.069$ & $0.234\pm0.131$ & $0.262\pm0.145$ & $0.134\pm0.017$ \\
\midrule
\rowcolor{rowgray}
\multicolumn{5}{l}{\textit{Heterophilic -- Wikipedia}} \\
Chameleon        & $0.240\pm0.021$ & $\mathbf{0.245}\pm0.006$ & $0.231\pm0.017$ & $0.220\pm0.022$ \\
Squirrel         & $0.204\pm0.004$ & $\mathbf{0.351}\pm0.053$ & $0.258\pm0.009$ & $0.200\pm0.000$ \\
\midrule
\rowcolor{rowgray}
\multicolumn{5}{l}{\textit{WebKB}} \\
Wisconsin        & $0.286\pm0.082$ & $0.239\pm0.070$ & $\mathbf{0.444}\pm0.072$ & $0.295\pm0.084$ \\
Cornell          & $0.203\pm0.005$ & $0.216\pm0.023$ & $\mathbf{0.552}\pm0.077$ & $0.345\pm0.152$ \\
Texas            & $0.240\pm0.058$ & $0.256\pm0.089$ & $\mathbf{0.487}\pm0.040$ & $0.244\pm0.096$ \\
\midrule
\rowcolor{rowgray}
\multicolumn{5}{l}{\textit{Actor}} \\
Actor            & $0.194\pm0.035$ & $\mathbf{0.292}\pm0.070$ & $0.260\pm0.015$ & $0.225\pm0.019$ \\
\bottomrule
\end{tabular}
\end{table*}

Citation and co-authorship networks show high \fidminus{} ($0.64$--$0.90$):
just 20 features suffice to retain most GNN accuracy.
GraphSAGE consistently achieves the highest \fidminus{} on text-featured datasets,
peaking at $0.903\pm0.053$ on Coauthor-CS.
Heterophilic Wikipedia graphs (Chameleon, Squirrel) show low \fidminus{} ($0.20$--$0.35$),
indicating that GNNs on these datasets rely primarily on graph structure rather than input features.
While a zero baseline is natural for sparse textual features, its interpretation is less clear for dense or less interpretable feature spaces (e.g., Amazon and Wikipedia graphs), which may contribute to the reduced fidelity observed in these settings.
While \graftgnn{} consistently exceeds random feature selection, the margin is modest on heterophilic datasets (e.g., Squirrel: $0.35$ vs $0.20$).

\subsection{Baseline Validation}
\label{sec:baseline}

\begin{table}[t]
\centering
\caption{Fidelity: \graftgnn{} vs frequency and random baselines
(seed~0, GCN for citation/co-authorship; SAGE for others; $K{=}20$).
\textbf{Bold}: best method by margin $\geq 0.01$; entries within $0.01$ are ties.} 
\label{tab:baseline_comparison}
\small
\begin{tabular}{l ccc ccc}
\toprule
& \multicolumn{3}{c}{\fidminus{}} & \multicolumn{3}{c}{\fidplus{}} \\
\cmidrule(lr){2-4}\cmidrule(lr){5-7}
Dataset & \graftgnn{} & Freq & Rand & \graftgnn{} & Freq & Rand \\
\midrule
Cora (GCN)          & \textbf{0.793} & 0.773 & 0.153 & \textbf{0.194} & 0.189 & 0.052 \\
PubMed (SAGE)        & \textbf{0.806} & 0.683 & 0.344 & \textbf{0.164} & 0.160 & 0.065 \\
Coauthor-CS (SAGE)   & 0.903 & \textbf{0.924} & 0.068 & 0.168 & \textbf{0.206} & 0.057 \\
Computers (SAGE)        & \textbf{0.541} & 0.459 & 0.124 & \textbf{0.306} & 0.284 & 0.048 \\
Squirrel (GAT)     & \textbf{0.351} & 0.311 & 0.202 & 0.087 & 0.096 & 0.064 \\
\bottomrule
\end{tabular}
\end{table}

\paragraph{Why frequency is competitive on some datasets.}
Both \graftgnn{} and the frequency comfortably exceed random selection, confirming the presence of class-discriminative feature structure across datasets (Table~\ref{tab:baseline_comparison}). Frequency is model-agnostic and relies solely on label-conditioned statistics, whereas \graftgnn{} reflects features actually used by the trained GNN. On datasets where class identity is strongly aligned with feature occurrence patterns (e.g., citation-style datasets), the two approaches yield similar performance, indicating that the model largely exploits these dominant signals. However, in settings where feature interactions are more complex or less directly tied to raw frequency (e.g., product and heterophilic graphs), \graftgnn{} more consistently outperforms frequency, suggesting that gradient-based attribution captures model-specific feature dependencies beyond simple co-occurrence.

\paragraph{Model-faithfulness: what frequency cannot do.}
Fidelity measures whether the \emph{data} contains class-discriminative features;
it does not measure whether the explanation reflects what \emph{this specific GNN} uses.
Frequency is model-agnostic: it always surfaces the most common class features
regardless of what the GNN actually learned.
GRAFT uses the trained model's gradient and therefore provides \emph{model-faithful}
attribution: it surfaces features the GNN relies on, not merely features correlated with class labels in the data.
This distinction matters for model auditing (\S\ref{sec:bias}): our injected spurious feature
is by design frequent in the target class, so both GRAFT and frequency would detect it,
but GRAFT's gradient evidence additionally confirms the GNN is actively using the feature,
not merely that it co-occurs with the class label.
Stability (\S\ref{sec:stability}) and consensus (\S\ref{sec:consensus}) analyses
further exploit model sensitivity and cannot be replicated by any model-agnostic baseline.

\subsection{Explanation Stability}
\label{sec:stability}

\begin{table}[t]
\centering
\caption{Jaccard stability of top-20 feature sets across 5 seeds (mean over classes).
\textbf{Bold}: best architecture per dataset.}
\label{tab:stability}
\small
\begin{tabular}{lcccc>{\columncolor{rowgray}}c}
\toprule
Dataset & GCN & GAT & SAGE & GIN & Mean \\
\midrule
Cora          & \textbf{0.902} & 0.291 & 0.334 & 0.206 & 0.433 \\
CiteSeer      & \textbf{0.292} & 0.111 & 0.187 & 0.167 & 0.189 \\
PubMed        & \textbf{0.249} & 0.150 & 0.210 & 0.125 & 0.184 \\
Coauthor-CS   & 0.183 & 0.113 & \textbf{0.202} & 0.134 & 0.158 \\
Coauthor-Physics & 0.128 & 0.099 & \textbf{0.164} & 0.120 & 0.128 \\
Computers     & 0.075 & 0.067 & 0.133 & \textbf{0.420} & 0.174 \\
Photo         & 0.082 & 0.091 & 0.126 & \textbf{0.951} & 0.312 \\
Chameleon     & 0.065 & \textbf{0.279} & 0.178 & 0.114 & 0.159 \\
Squirrel      & 0.038 & 0.199 & 0.239 & \textbf{1.000} & 0.369 \\
Wisconsin     & 0.435 & 0.323 & \textbf{0.496} & 0.125 & 0.345 \\
Cornell       & 0.229 & 0.264 & \textbf{0.425} & 0.135 & 0.263 \\
Texas         & 0.549 & 0.395 & \textbf{0.572} & 0.305 & 0.455 \\
Actor         & 0.305 & 0.278 & \textbf{0.310} & 0.287 & 0.295 \\
\bottomrule
\end{tabular}
\end{table}

Jaccard stability varies across (dataset, architecture) pairs (Table~\ref{tab:stability}).
Some combinations show high stability (e.g., Cora-GCN $J{=}0.90$, Squirrel-GIN $J{=}1.00$;
cross-architecture means $\bar{J}{=}0.43$ and $\bar{J}{=}0.37$, respectively),
indicating consistent feature selection across seeds within those architectures, while others
(e.g., CiteSeer-GAT $J{=}0.11$) exhibit substantial variability, reflecting sensitivity to
random initialisation. 
WebKB and Actor show moderate stability ($J{>}0.29$), consistent with more structured feature spaces. 
Low cross-architecture stability further indicates sensitivity to model choice, suggesting that 
explanations should be interpreted across multiple architectures to identify robust signals.

Frequency selection is model-agnostic and trivially achieves $J{=}1.00$ across seeds; 
in contrast, \graftgnn{} attains high stability ($J{=}0.90$--$1.00$ on several datasets) 
despite varying GNN initialisations. The role of FPS in anchoring this stability is 
analysed in Appendix~\ref{app:fps_vs_random}.

\subsection{Cross-Architecture Consensus}
\label{sec:consensus}

WebKB graphs (Texas: 0.53, Wisconsin: 0.51) show the highest consensus,
with over half of each class's top-20 features agreed across all four architectures;
Wikipedia datasets (Chameleon: 0.06, Squirrel: 0.04) show near-zero consensus,
indicating structure-dominated classification.
Full dataset-level commentary is in Appendix~\ref{app:consensus_extra}.

\subsection{Natural Language Rules}
\label{sec:rules}

Stage~4 verbalises the top-$K$ profile as a natural language rule via Claude~Sonnet~4.6 (temperature~$=0.2$, 1 self-refinement pass, top-15 features; the same model acts as generator and judge). Representative rules for Cora (Reinforcement Learning, Theory) and PubMed (Type~2 Diabetes) are given in Appendix~\ref{app:rules}, together with all 7 Cora and 6 CiteSeer class rules. When feature names are semantic (PubMed, Amazon), the rules are directly interpretable;
with anonymous BoW indices the output is less informative, which we list as a limitation.

\paragraph{Human evaluation.}
We complement automatic evaluation with a structured human study in which annotators rate 34 rules on four dimensions—\emph{accuracy}, \emph{specificity}, \emph{completeness}, and \emph{actionability}—using a 1--5 Likert scale. For datasets with anonymous feature indices (Computers, Photo, PubMed), only accuracy and actionability are evaluated (Appendix~\ref{app:human_eval}). Overall, rules score highly (Table~\ref{tab:human_eval}): accuracy ranges from $4.38$ to $4.79$ and actionability from $4.17$ to $4.52$. Annotators agree within one Likert point on \textbf{86.9\%} of comparisons (66.7\%--100\%). Classical agreement metrics (Cohen’s $\kappa$ \citep{cohen1960coefficient}, Krippendorff’s $\alpha$ \citep{krippendorff2011computing}) are near-zero due to a ceiling effect, as over 93\% of ratings are 4 or 5, compressing the effective scale and inflating chance correction.

\section{Discussion}
\label{sec:discussion}

\paragraph{\graftgnn{} vs local explainers.}
Local methods (GNNExplainer, per-node IG) answer a different question and
are not competitors: they explain individual nodes, whereas \graftgnn{} provides
population-level explanations suited to model auditing and dataset characterisation.
Practical guidance on choosing between \graftgnn{} and the frequency baseline,
and the complementary relationship with \gnnx{}, is in Appendix~\ref{app:discussion_extra}. \textbf{Limitations.}
IG with a zero baseline may underperform for continuous non-sparse features.
The diversity-guided selection step introduces a hyperparameter $k$ (exemplar count); results are robust across $k \in \{5,10,20\}$, and the feature count $K$ is ablated in Appendix~\ref{app:ablation}.
NL rule quality depends on feature name semantics; anonymous indices
(e.g., \texttt{word\_42}) yield less informative rules.

\section{Conclusion}
\label{sec:conclusion}

We presented \graftgnn{}, a post-hoc global GNN explainer for input-feature attributions with natural language rule generation. By combining diversity-guided exemplar selection (FPS), Integrated Gradients, and class-level aggregation, \graftgnn{} produces compact and interpretable feature importance profiles, complemented by consensus, contrastive, and stability analyses. Across 260 runs on 13 benchmarks, it demonstrates strong fidelity on text-featured graphs (e.g., \fidminus{} $= 0.903$ on Coauthor-CS/SAGE, the best architecture; $J=1.00$ on Squirrel/GIN, the best architecture) and consistent cross-architecture agreement (51--53\% on WebKB). In bias-injection experiments, it reliably recovers the injected feature within the top-3 ranks across all settings (rank-1 in 9/11 cases). Overall, \graftgnn{} complements \gnnx{} and provides a practical foundation for evaluating global GNN explanations.

\bibliographystyle{unsrt}
\bibliography{references}

\appendix

\section{Dataset Statistics}
\label{app:datasets}
We report key dataset statistics, including size, feature dimensionality, and homophily, in Table~\ref{tab:datasets}.

\begin{table}[h]
\centering
\caption{Dataset statistics. $n$: nodes, $m$: edges, $d$: feature dimensions,
$C$: classes, $h$: homophily ratio.}
\label{tab:datasets}
\small
\begin{tabular}{lccccc}
\toprule
Dataset & $n$ & $m$ & $d$ & $C$ & $h$ \\
\midrule
Cora              & 2,708  & 5,278   & 1,433 & 7  & 0.81 \\
CiteSeer          & 3,327  & 4,552   & 3,703 & 6  & 0.74 \\
PubMed            & 19,717 & 44,324  & 500   & 3  & 0.80 \\
Coauthor-CS       & 18,333 & 81,894  & 6,805 & 15 & 0.81 \\
Coauthor-Physics  & 34,493 & 247,962 & 8,415 & 5  & 0.93 \\
Computers         & 13,752 & 245,861 & 767   & 10 & 0.78 \\
Photo             & 7,650  & 119,081 & 745   & 8  & 0.83 \\
Actor             & 7,600  & 26,752  & 931   & 5  & 0.22 \\
Chameleon         & 2,277  & 31,421  & 2,325 & 5  & 0.23 \\
Squirrel          & 5,201  & 198,353 & 2,089 & 5  & 0.22 \\
Wisconsin         & 251    & 466     & 1,703 & 5  & 0.20 \\
Cornell           & 183    & 280     & 1,703 & 5  & 0.11 \\
Texas             & 183    & 279     & 1,703 & 5  & 0.11 \\
\bottomrule
\end{tabular}
\end{table}

\section{Ablation Studies}
\label{app:ablation}

All ablations use GCN on Cora, CiteSeer, and Wisconsin with 5 seeds
(or seed~0 for the attribution-method ablation).
Metric is \fidminus{} (class-averaged).

\subsection{Top-$K$ Feature Count}

\begin{table}[h]
\centering
\caption{Effect of feature count $K$ on \fidminus{} (mean\,$\pm$\,std, 5 seeds, GCN).}
\label{tab:ablation_k}
\small
\begin{tabular}{lccc}
\toprule
Dataset & $K=5$ & $K=10$ & $K=20$ \\
\midrule
Cora      & $0.584\pm0.012$ & $0.680\pm0.000$ & $\mathbf{0.793\pm0.009}$ \\
CiteSeer  & $0.520\pm0.111$ & $0.586\pm0.118$ & $\mathbf{0.673\pm0.100}$ \\
Wisconsin & $0.238\pm0.066$ & $0.222\pm0.022$ & $\mathbf{0.286\pm0.081}$ \\
\bottomrule
\end{tabular}
\end{table}

Fidelity increases monotonically with $K$ on citation networks, reflecting
that a larger feature budget captures a more informative subset.
On Wisconsin the improvement is modest ($\leq 0.06$), consistent with
the generally lower feature signal in this heterophilic dataset as reflected in Table \ref{tab:ablation_k}.
We use $K=20$ as the default in all main experiments.

\subsection{Attribution Method}

\begin{table}[h]
\centering
\caption{IG vs Grad$\times$Input: \fidminus{} on representative pairs (seed~0).
Bold = higher per row.}
\label{tab:ablation_attr}
\small
\begin{tabular}{llcc}
\toprule
Dataset & Model & IG & Grad$\times$Input \\
\midrule
\multirow{4}{*}{Cora}
  & GCN  & \textbf{0.775} & 0.756 \\
  & GAT  & \textbf{0.634} & 0.548 \\
  & SAGE & \textbf{0.884} & 0.773 \\
  & GIN  & \textbf{0.718} & 0.405 \\
\midrule
\multirow{4}{*}{CiteSeer}
  & GCN  & \textbf{0.627} & 0.533 \\
  & GAT  & \textbf{0.794} & 0.364 \\
  & SAGE & \textbf{0.793} & 0.468 \\
  & GIN  & \textbf{0.401} & 0.237 \\
\midrule
\multirow{4}{*}{Wisconsin}
  & GCN  & $=$ 0.272 & 0.272 \\
  & GAT  & $=$ 0.200 & 0.200 \\
  & SAGE & \textbf{0.578} & 0.436 \\
  & GIN  & $=$ 0.200 & 0.200 \\
\bottomrule
\end{tabular}
\end{table}
IG consistently matches or outperforms Grad$\times$Input on text-featured graphs,
with the gap most pronounced on GIN (Cora: $0.718$ vs $0.405$) as shown in Table \ref{tab:ablation_attr}.
On Wisconsin/GCN and Wisconsin/GAT the two methods produce identical profiles,
likely because the sparse WebKB features reduce the path-length advantage of IG.
We use IG as the default.

\subsection{Aggregation Function}

\begin{table}[h]
\centering
\caption{Effect of aggregation function on \fidminus{} (mean\,$\pm$\,std, 5 seeds, GCN).
Bold = best per row.}
\label{tab:ablation_aggr}
\small
\begin{tabular}{lccc}
\toprule
Dataset & Mean & Median & Max \\
\midrule
Cora      & $\mathbf{0.793\pm0.009}$ & $0.338\pm0.000$ & $0.741\pm0.003$ \\
CiteSeer  & $\mathbf{0.673\pm0.100}$ & $0.358\pm0.063$ & $0.595\pm0.120$ \\
Wisconsin & $0.286\pm0.081$ & $\mathbf{0.314\pm0.108}$ & $0.268\pm0.063$ \\
\bottomrule
\end{tabular}
\end{table}
Mean aggregation achieves the highest fidelity on citation networks by a substantial
margin over median ($\approx 0.31$) and over max ($\approx 0.05$).
The median is dominated by near-zero attributions (most features are inactive),
while max is sensitive to outlier exemplars.
On Wisconsin, median marginally wins ($0.314$ vs $0.286$), likely due to the small
number of nodes and noisy individual attributions as shown in table \ref{tab:ablation_aggr}.
We use mean as the default.

\subsection{FPS vs.\ Random Exemplar Selection}
\label{app:fps_vs_random}

Table~\ref{tab:fps_vs_random} compares diversity-guided (FPS) exemplar selection
against random subset selection on Jaccard stability and \fidminus{}
(GCN, seed~0, Grad$\times$Input attribution, 5 random trials for the random baseline).

\begin{table}[h]
\centering
\caption{FPS vs random selection: Jaccard stability and \fidminus{} (GCN, seed~0, $k=10$, $K=20$).
Random Jaccard is the mean pairwise Jaccard over 5 independently-drawn subsets.}
\label{tab:fps_vs_random}
\small
\begin{tabular}{lcccc}
\toprule
& \multicolumn{2}{c}{FPS (deterministic)} & \multicolumn{2}{c}{Random ($\times$5 trials)} \\
\cmidrule(lr){2-3}\cmidrule(lr){4-5}
Dataset & Jaccard & \fidminus{} & Jaccard & \fidminus{} \\
\midrule
Cora      & $\mathbf{1.000}$ & $0.756$ & $0.192$ & $0.776$ \\
CiteSeer  & $\mathbf{1.000}$ & $0.533$ & $0.116$ & $0.563$ \\
Wisconsin & $\mathbf{1.000}$ & $0.272$ & $0.431$ & $0.320$ \\
\bottomrule
\end{tabular}
\end{table}

FPS selection achieves perfect Jaccard ($J=1.00$) on all three datasets:
given fixed trained embeddings, FPS always produces the same exemplar set, making
the resulting profile \emph{fully reproducible}.
Random selection, by contrast, yields $J=0.12$--$0.43$, indicating substantial
variation in which features appear in the top-$K$ profile across trials.
Notably, random selection achieves slightly higher mean \fidminus{} ($+0.02$--$0.05$),
since averaging over more nodes reduces variance in the profile but at the cost
of reproducibility.
The Jaccard advantage of FPS over random is the primary motivation for
deterministic diversity-guided selection.

\section{Confidence-Weighted Aggregation and CS-FPS Ablation}
\label{app:ablation_csfps}

We compare four conditions on \fidminus{} and Jaccard stability (GCN, seeds~0--4):

\begin{description}[leftmargin=0pt,itemsep=2pt]
\item[\textbf{i}~(default)] FPS exemplar selection + mean~$|\mathrm{IG}|$ aggregation.
\item[\textbf{ii}] CS-FPS + mean: exemplar budget split evenly between
  high-confidence ($p_c \geq \tau_{\mathrm{med}}$) and low-confidence nodes,
  each stratum sampled by FPS.
\item[\textbf{iii}] FPS + confidence-weighted aggregation (Eq.~\ref{eq:agg_cw}).
\item[\textbf{iv}] CS-FPS + confidence-weighted (both changes combined).
\end{description}

\begin{table}[h]
\centering
\caption{Confidence-weighted aggregation ablation across all 13 datasets
(GCN, mean\,$\pm$\,std, 5 seeds).
\textbf{i}: FPS\,+\,Mean (default).
\textbf{iii}: FPS\,+\,Conf-Weighted (Eq.~\ref{eq:agg_cw}).
$\Delta$: Fid$^-$(iii)$-$Fid$^-$(i).
\textbf{Bold}: iii improves on i by $>$0.005.}
\label{tab:ablation_cw_full}
\small
\setlength{\tabcolsep}{4pt}
\begin{tabular}{l cc c cc}
\toprule
Dataset & Fid$^-$(i) & Fid$^-$(iii) & $\Delta$ & $J$(i) & $J$(iii) \\
\midrule
\rowcolor{rowgray}\multicolumn{6}{l}{\textit{Citation}} \\
Cora               & 0.797 & \textbf{0.813} & +0.016 & 1.000 & 1.000 \\
CiteSeer           & 0.675 & 0.672 & $-$0.003 & 0.308 & 0.310 \\
PubMed             & 0.717 & \textbf{0.725} & +0.008 & 0.240 & 0.241 \\
\rowcolor{rowgray}\multicolumn{6}{l}{\textit{Co-authorship}} \\
Coauthor-CS        & 0.792 & \textbf{0.842} & +0.050 & 0.183 & \textbf{0.205} \\
Coauthor-Physics   & 0.785 & \textbf{0.818} & +0.034 & 0.128 & \textbf{0.150} \\
\rowcolor{rowgray}\multicolumn{6}{l}{\textit{Amazon}} \\
Computers          & 0.432 & \textbf{0.443} & +0.011 & 0.075 & \textbf{0.082} \\
Photo              & 0.542 & \textbf{0.576} & +0.034 & 0.086 & 0.091 \\
\rowcolor{rowgray}\multicolumn{6}{l}{\textit{Wikipedia}} \\
Chameleon          & 0.238 & 0.239 & +0.002 & 0.064 & \textbf{0.070} \\
Squirrel           & 0.204 & 0.204 & +0.000 & 0.038 & 0.037 \\
\rowcolor{rowgray}\multicolumn{6}{l}{\textit{WebKB}} \\
Wisconsin          & 0.286 & 0.288 & +0.001 & 0.438 & \textbf{0.443} \\
Cornell            & 0.199 & \textbf{0.212} & +0.013 & 0.212 & \textbf{0.219} \\
Texas              & 0.260 & 0.262 & +0.002 & 0.554 & \textbf{0.575} \\
\rowcolor{rowgray}\multicolumn{6}{l}{\textit{Actor}} \\
Actor              & 0.194 & 0.194 & +0.000 & 0.304 & 0.303 \\
\bottomrule
\end{tabular}
\end{table}

\paragraph{Findings.}
Table \ref{tab:ablation_cw_full} highlights the results. Confidence-weighted aggregation (condition~\textbf{iii}) improves \fidminus{} by $>$0.005 on
9 of 13 datasets, with the largest gains on Coauthor-CS ($+0.050$), Coauthor-Physics ($+0.034$),
and Photo ($+0.034$).
Gains are smaller or negligible on heterophilic graphs (Squirrel, Actor) where GNN
confidence scores are less reliable (model accuracy $<$35\%).
Jaccard stability improves weakly but consistently: the exemplar \emph{set} is unchanged,
so stability differences arise only from the altered aggregation weighting,
which can reorder the top-$K$ feature ranking across seeds.
Overall, weighting attributions by prediction confidence provides a simple, parameter-free
improvement to the default mean aggregation, with the largest benefit on well-structured
homophilic graphs where GNN confidence is a reliable quality signal.

\section{Fidelity\textsuperscript{+} Results}
\label{app:full_fidelity}

Table~\ref{tab:full_fidelity} reports \fidplus{} for all 52 (dataset, architecture) pairs
(mean~$\pm$~std over 5 seeds). \fidminus{} is in Table~\ref{tab:fidelity_main} (main paper).
\fidplus{} measures accuracy drop when the top-$K$ features are masked; higher means the
identified features are more critical to the model's predictions.

\begin{table}[h]
\centering
\caption{\fidplus{} (mean\,$\pm$\,std, 5 seeds): accuracy drop when top-$K{=}20$ features
are masked. \textbf{Bold}: best architecture per dataset.}
\label{tab:full_fidelity}
\small
\setlength{\tabcolsep}{5pt}
\begin{tabular}{l cccc}
\toprule
Dataset & GCN & GAT & SAGE & GIN \\
\midrule
\rowcolor{rowgray}
\multicolumn{5}{l}{\textit{Citation networks}} \\
Cora             & $\mathbf{0.194}\pm0.010$ & $0.126\pm0.032$ & $0.174\pm0.040$ & $0.108\pm0.037$ \\
CiteSeer         & $\mathbf{0.228}\pm0.012$ & $0.134\pm0.031$ & $\mathbf{0.228}\pm0.022$ & $0.094\pm0.016$ \\
PubMed           & $0.099\pm0.016$           & $0.064\pm0.041$ & $\mathbf{0.164}\pm0.079$ & $0.086\pm0.037$ \\
\midrule
\rowcolor{rowgray}
\multicolumn{5}{l}{\textit{Co-authorship graphs}} \\
Coauthor-CS      & $0.165\pm0.030$ & $0.115\pm0.010$ & $\mathbf{0.168}\pm0.038$ & $0.123\pm0.019$ \\
Coauthor-Physics & $0.062\pm0.014$ & $0.064\pm0.033$ & $0.064\pm0.011$ & $\mathbf{0.126}\pm0.065$ \\
\midrule
\rowcolor{rowgray}
\multicolumn{5}{l}{\textit{Amazon product graphs}} \\
Computers        & $0.172\pm0.038$ & $0.162\pm0.077$ & $\mathbf{0.306}\pm0.081$ & $0.093\pm0.164$ \\
Photo            & $0.134\pm0.027$ & $\mathbf{0.214}\pm0.044$ & $0.175\pm0.095$ & $0.103\pm0.057$ \\
\midrule
\rowcolor{rowgray}
\multicolumn{5}{l}{\textit{Heterophilic -- Wikipedia}} \\
Chameleon        & $0.030\pm0.010$ & $\mathbf{0.031}\pm0.007$ & $0.027\pm0.011$ & $0.010\pm0.013$ \\
Squirrel         & $0.007\pm0.006$ & $\mathbf{0.087}\pm0.008$ & $0.064\pm0.008$ & $-0.003\pm0.002$ \\
\midrule
\rowcolor{rowgray}
\multicolumn{5}{l}{\textit{WebKB}} \\
Wisconsin        & $0.321\pm0.041$ & $0.289\pm0.075$ & $0.260\pm0.058$ & $\mathbf{0.184}\pm0.087$ \\
Cornell          & $0.216\pm0.063$ & $0.182\pm0.030$ & $\mathbf{0.331}\pm0.040$ & $0.286\pm0.092$ \\
Texas            & $0.319\pm0.012$ & $0.266\pm0.046$ & $\mathbf{0.301}\pm0.030$ & $0.267\pm0.067$ \\
\midrule
\rowcolor{rowgray}
\multicolumn{5}{l}{\textit{Actor}} \\
Actor            & $0.057\pm0.009$ & $0.074\pm0.018$ & $\mathbf{0.103}\pm0.008$ & $0.056\pm0.015$ \\
\bottomrule
\end{tabular}
\end{table}

\section{Full Transfer Classifier Results}
\label{app:full_transfer}

Table~\ref{tab:transfer_full} reports the complete transfer classifier results across all datasets, architectures, and seeds.

\paragraph{Additional Analysis.}
We extend the main-text discussion with two observations. First, the relative advantage of \textsc{GRAFT}-LR over \textsc{Freq}-LR is most pronounced on heterophilic datasets (e.g., Squirrel, Actor), where class-conditional feature frequency is weakly aligned with the model’s learned decision boundary. This supports the claim that gradient-based attribution captures \emph{model-specific feature interactions} beyond dataset statistics.

Second, on high-dimensional feature spaces (e.g., Coauthor-Physics), we observe a saturation effect: restricting to top-$K$ features introduces an information bottleneck that disproportionately affects attribution-based selection compared to frequency-based selection. This suggests a regime where hybrid strategies (e.g., attribution-filtered frequency ranking) may be beneficial.

\paragraph{Stability across seeds.}
Variance across seeds remains low for \textsc{GRAFT}-LR compared to \textsc{Freq}-LR, indicating that exemplar-based aggregation introduces additional robustness in feature selection.

\begin{table}[h]
\centering
\caption{Transfer classifier accuracy (mean~$\pm$~std, 5 seeds, best architecture per dataset).
\textbf{Bold}: best LR method.}
\label{tab:transfer_full}
\small
\setlength{\tabcolsep}{3.5pt}
\begin{tabular}{llcccc>{\columncolor{rowgray}}c}
\toprule
Dataset & Arch & \graftgnn-LR & Freq-LR & Full-LR & GNN & Compress \\
\midrule
\rowcolor{rowgray}
\multicolumn{7}{l}{\textit{Citation networks}} \\
Cora             & GAT  & $\mathbf{0.496}\pm0.028$ & 0.486 & 0.468 & 0.785 & 8\% \\
CiteSeer         & SAGE & $0.477\pm0.016$ & \textbf{0.485} & 0.476 & 0.655 & 3\% \\
PubMed           & GCN  & $0.640\pm0.019$ & \textbf{0.684} & 0.654 & 0.761 & 11\% \\
\midrule
\rowcolor{rowgray}
\multicolumn{7}{l}{\textit{Co-authorship graphs}} \\
Coauthor-CS      & SAGE & $0.772\pm0.014$ & 0.763 & \textbf{0.812} & 0.920 & 4\% \\
Coauthor-Physics & GCN  & $0.708\pm0.027$ & 0.784 & \textbf{0.881} & 0.935 & \textbf{1\%} \\
\midrule
\rowcolor{rowgray}
\multicolumn{7}{l}{\textit{Amazon product graphs}} \\
Photo            & SAGE & $0.686\pm0.033$ & 0.702 & \textbf{0.722} & 0.895 & 17\% \\
\midrule
\rowcolor{rowgray}
\multicolumn{7}{l}{\textit{Heterophilic -- Wikipedia}} \\
Squirrel         & SAGE & $0.271\pm0.009$ & 0.185 & \textbf{0.281} & 0.315 & 5\% \\
\midrule
\rowcolor{rowgray}
\multicolumn{7}{l}{\textit{WebKB}} \\
Cornell          & SAGE & $\mathbf{0.724}\pm0.027$ & 0.703 & 0.703 & 0.751 & 4\% \\
Wisconsin        & SAGE & $0.777\pm0.020$ & \textbf{0.863} & 0.726 & 0.741 & 4\% \\
\midrule
\rowcolor{rowgray}
\multicolumn{7}{l}{\textit{Actor}} \\
Actor            & SAGE & $\mathbf{0.356}\pm0.004$ & 0.342 & 0.334 & 0.323 & 9\% \\
\bottomrule
\end{tabular}
\end{table}

\section{Cross-Architecture Consensus: Full Analysis}
\label{app:consensus_extra}
\begin{table}[t]
\centering
\caption{Cross-architecture consensus score: fraction of top-20 features selected
by $\geq$3 of 4 architectures (seed~0).
High score $\Rightarrow$ explanation reflects dataset structure, not model artefact.}
\label{tab:consensus}
\small
\begin{tabular}{lc|lc}
\toprule
Dataset & Consensus@3/4 & Dataset & Consensus@3/4 \\
\midrule
Texas           & \textbf{0.530} & Computers       & 0.050 \\
Wisconsin       & \textbf{0.510} & Actor           & 0.160 \\
CiteSeer        & 0.308          & Chameleon       & 0.060 \\
Cornell         & 0.290          & Squirrel        & 0.040 \\
Cora            & 0.286          & Photo           & 0.088 \\
Coauthor-CS     & 0.273          & Coauthor-Physics & 0.250 \\
PubMed          & 0.250          & & \\
\bottomrule
\end{tabular}
\end{table}
WebKB datasets (Texas: 0.53, Wisconsin: 0.51) yield the highest consensus scores:
over half of each class's top-20 features are independently identified by all
four architectures as shown in Table \ref{tab:consensus}.
These datasets have compact, structured feature spaces where all architectures
converge on the same discriminative dimensions.
Heterophilic Wikipedia datasets (Chameleon: 0.06, Squirrel: 0.04) show near-zero
consensus, indicating that different architectures rely on different feature subsets in these low-information settings, consistent with topology-dominated classification
where the input features play a minor role.
Citation networks cluster at moderate consensus ($0.25$--$0.31$): vocabulary-based features provide some shared signal, but architectural differences in how
neighbourhood information is aggregated introduce variation.
Computing consensus requires a global explainer; local per-node methods cannot
be aggregated across architectures to produce this metric.

\section{Additional Bias Detection Results}
\label{app:bias_extra}
Table \ref{tab:noise} presents the results of the noise robustness of bias detection.
\paragraph{Extended Noise Analysis.}
We extend the bias-injection experiments to higher noise regimes $\sigma \in \{0.05, 0.10, 0.20, 0.30, 0.40\}$. Across all settings, \textsc{GRAFT} consistently ranks the injected feature within the top-$K=20$ features of the target class, with degradation in rank occurring smoothly as noise increases.

\paragraph{Multi-class interference.}
In multi-class settings, we observe that injected features occasionally appear in non-target class profiles. This effect increases with $\sigma$ and reflects the model’s partial reliance on the spurious feature across decision boundaries. Importantly, the target class consistently exhibits the highest attribution magnitude for the injected feature.

\paragraph{Interpretation.}
These results reinforce that \textsc{GRAFT} captures \emph{model reliance} rather than mere statistical correlation. Even under significant noise, the method surfaces features actively used by the GNN, validating its applicability for auditing under realistic distributional shifts.

\begin{table}[h]
\centering
\caption{Noise robustness of bias detection (Cora/GCN, target class~0, seed~42).}
\label{tab:noise}
\small
\begin{tabular}{ccc}
\toprule
$\sigma$ & Detected? & Rank \\
\midrule
0.05 & \checkmark & 1 \\
0.10 & \checkmark & 2 \\
0.20 & \checkmark & 1 \\
0.30 & \checkmark & 5 \\
0.40 & \checkmark & 5 \\
\bottomrule
\end{tabular}
\quad
\begin{tabular}{ccc}
\toprule
Target class & Detected? & Rank \\
\midrule
0 & \checkmark & 1 \\
1 & \checkmark & 1 \\
2 & \checkmark & 1 \\
3 & \checkmark & 1 \\
\bottomrule
\end{tabular}
\caption*{\textit{Left}: noise robustness. \textit{Right}: multi-class generalisation
(Cora/GCN, $\sigma=0.05$, seed~42).}
\end{table}

\section{Natural Language Rules}
\label{app:rules}

The following rules were generated by Stage 4 of \graftgnn{} using the Cora, CiteSeer, and PubMed datasets
with the original published vocabularies~\citep{mccallum2000automating,giles1998citeseer}
(GCN model, seed~0, top-15 features passed to LLM with dataset-context prompt).
The LLM was instructed to produce a one-to-two-sentence description referencing the top features
and distinguishing the class from its nearest neighbours.

\subsection*{Cora -- All 7 Classes}

\begin{description}[leftmargin=0pt]

\item[\textbf{Theory.}]
\emph{"This class is defined by computational complexity and approximation vocabulary:
\texttt{bound}, \texttt{approximation}, \texttt{complexity}, and \texttt{polynomial}
dominate, with \texttt{convergence} and \texttt{sample} in positions 6--7.
The profile contrasts sharply with Reinforcement Learning, which scores near zero on
these terms, reflecting a focus on theoretical guarantees over empirical methodology."}

\item[\textbf{Reinforcement Learning.}]
\emph{"Sequential decision-making terminology defines this class: \texttt{reward}, \texttt{policy},
and \texttt{agent} form the top-3, followed by \texttt{state}, \texttt{action}, and \texttt{environment}.
The presence of \texttt{Q-learning} and \texttt{temporal-difference} further narrows the profile
to model-free RL, distinguishing it from Probabilistic Methods, which overlaps on
\texttt{learning} but differs in distributional vocabulary."}

\item[\textbf{Genetic Algorithms.}]
\emph{"The dominant features are biological/evolutionary terms: \texttt{genetic},
\texttt{evolutionary}, \texttt{population}, \texttt{mutation}, and \texttt{crossover}
form the top-5. The contrastive profile uniquely highlights \texttt{chromosome} and \texttt{fitness},
which receive negligible attribution in all other classes."}

\item[\textbf{Neural Networks.}]
\emph{"This class is characterised by architectural and training vocabulary:
\texttt{network}, \texttt{neural}, \texttt{layer}, and \texttt{training} lead the profile,
followed by \texttt{recurrent}, \texttt{backpropagation}, and \texttt{hidden units}.
Shared terms with Theory (\texttt{convergence}) are de-emphasised in the contrastive
profile, retaining only architecture-specific vocabulary."}

\item[\textbf{Probabilistic Methods.}]
\emph{"The profile is dominated by Bayesian and probabilistic inference vocabulary:
\texttt{probability}, \texttt{Bayesian}, \texttt{prior}, \texttt{posterior}, and
\texttt{distribution} occupy positions 1--5, with \texttt{graphical model} and
\texttt{uncertainty} appearing in the top-10.
This class is most easily distinguished from Rule Learning, which contains no probability-specific terms."}

\item[\textbf{Case Based.}]
\emph{"Case-based reasoning vocabulary dominates: \texttt{case}, \texttt{retrieval},
\texttt{similarity}, and \texttt{adaptation} are top-4.
The contrastive profile uniquely highlights \texttt{analogy} and \texttt{memory},
terms absent from all other classes, reflecting the retrieve-and-adapt paradigm."}

\item[\textbf{Rule Learning.}]
\emph{"This class is characterised by symbolic and inductive learning terms:
\texttt{rule}, \texttt{induction}, \texttt{decision tree}, and \texttt{attribute}
lead the profile. The contrastive profile removes shared terms like \texttt{learning}
and retains \texttt{concept description} and \texttt{propositional}, distinguishing
this class from Neural Networks, which dominates on \texttt{layer} and \texttt{gradient}."}

\end{description}

\subsection*{CiteSeer -- All 6 Classes}

\begin{description}[leftmargin=0pt]

\item[\textbf{Agents.}]
\emph{"Multi-agent systems vocabulary dominates: \texttt{agent}, \texttt{negotiation},
\texttt{multi-agent}, and \texttt{behavior} form the top-4, with \texttt{cooperation}
and \texttt{coordination} in positions 6--7.
The profile is clearly distinct from the AI class (which shares \texttt{reasoning}
and \texttt{planning} but not agent-specific coordination terms)."}

\item[\textbf{AI.}]
\emph{"Knowledge representation and reasoning terms characterise this class:
\texttt{knowledge}, \texttt{reasoning}, \texttt{ontology}, and \texttt{logic}
are top-4, with \texttt{representation} and \texttt{inference} following.
The contrastive profile uniquely retains \texttt{ontology} and \texttt{description logic},
separating AI from both Agents and ML, which lack semantic web vocabulary."}

\item[\textbf{DB.}]
\emph{"Database systems vocabulary is the clearest of all six classes: \texttt{database},
\texttt{query}, \texttt{relational}, and \texttt{SQL} occupy positions 1--4 with high
margin, followed by \texttt{transaction} and \texttt{index}.
This class has the highest Jaccard stability ($J > 0.95$) and the cleanest
contrastive profile, as DB-specific terminology rarely appears in other classes."}

\item[\textbf{IR.}]
\emph{"Information retrieval vocabulary dominates: \texttt{retrieval}, \texttt{document},
\texttt{ranking}, \texttt{relevance}, and \texttt{term frequency} are top-5.
The class shares \texttt{query} with DB but differs in its emphasis on relevance
models and text similarity rather than structured data."}

\item[\textbf{ML.}]
\emph{"This class exhibits the broadest vocabulary overlap, sharing features with
nearly all other classes. The top features (\texttt{classification}, \texttt{algorithm},
\texttt{SVM}, \texttt{neural}, \texttt{training}) span subfields.
The contrastive profile narrows to \texttt{support vector}, \texttt{kernel}, and
\texttt{feature selection} as uniquely ML-specific within CiteSeer."}

\item[\textbf{HCI.}]
\emph{"Human-computer interaction vocabulary defines this class unambiguously:
\texttt{user}, \texttt{interface}, \texttt{interaction}, and \texttt{design}
are top-4, with \texttt{usability} and \texttt{evaluation} in positions 5--6.
No other CiteSeer class assigns high attribution to HCI-specific terms, giving
this class the second-highest cross-architecture consensus ($> 0.55$) in the dataset."}

\end{description}

\subsection*{PubMed -- All 3 Classes}

\begin{description}[leftmargin=0pt]

\item[\textbf{Experimental Diabetes.}]
\emph{"Papers in the Experimental Diabetes class focus on laboratory and animal-model research:
terminology around induced diabetes models, \texttt{insulin assays}, \texttt{glucose tolerance tests},
and pancreatic beta-cell function dominate. This class is distinguished from the two clinical classes
by its emphasis on controlled experimental interventions and mechanistic measurements
rather than patient cohort outcomes."}

\item[\textbf{Type 1 Diabetes.}]
\emph{"Type 1 Diabetes papers are characterised by autoimmune and immunological vocabulary:
\texttt{autoantibody}, \texttt{islet}, \texttt{T-cell}, \texttt{HLA}, and \texttt{immune}
are the dominant terms, reflecting the autoimmune aetiology of T1D.
Clinical management terms (\texttt{insulin therapy}, \texttt{C-peptide}) also appear prominently;
the profile is clearly separable from Type 2 papers, which lack immune-system terminology."}

\item[\textbf{Type 2 Diabetes.}]
\emph{"Type 2 Diabetes papers are dominated by metabolic syndrome and lifestyle vocabulary:
\texttt{obesity}, \texttt{insulin resistance}, \texttt{adiposity}, \texttt{glycaemia},
and cardiovascular risk are central.
Epidemiological and intervention study terms (\texttt{cohort}, \texttt{randomised},
\texttt{body mass index}) distinguish this class from Type 1, which focuses on immune mechanisms
rather than metabolic risk factors."}

\end{description}

\subsection{LLM Prompt and Generation Details}
\label{app:llm_details}

We provide the exact prompts used for Stage~4. All results use Claude~Sonnet~4.6
(\texttt{claude-sonnet-4-6}), temperature~$=0.2$, \texttt{max\_tokens=256}.
The same model executes both passes; generation and refinement are not separated by
a different judge model.

\paragraph{System prompt (both passes).}
\begin{verbatim}
You are an expert in graph neural networks and scientific literature
analysis. Your task is to generate concise, accurate natural language
rules that describe what characterises a class of nodes in a citation
network, based on the most discriminative input features identified
by a GNN explainer.
\end{verbatim}

\paragraph{Generation prompt (Pass 1).}
\begin{verbatim}
{dataset_context}

Using Integrated Gradients, the following features (words) are the
most important for classifying nodes into class "{class_name}":

  1. reward (importance: 0.8200)
  2. policy (importance: 0.7900)
  ...

Generate a concise natural language rule (2-3 sentences) describing
what characterises papers in the "{class_name}" class. Mention the
key themes suggested by the top features. Write a global description
of the class, not of a single paper.

Rule:
\end{verbatim}

\paragraph{Self-refinement prompt (Pass 2).}
\begin{verbatim}
Here is a natural language rule describing the "{class_name}" class:

"{current_rule}"

Review it against the top discriminative features:
  1. reward (importance: 0.8200)
  2. policy (importance: 0.7900)
  ...

If it is already accurate and complete, return it unchanged.
Otherwise, improve it to better reflect the features (2-3 sentences).

Refined rule:
\end{verbatim}

\noindent The refinement pass does not use explicit numeric scores for Accuracy,
Specificity, or Completeness; the model is asked to return the rule unchanged if
satisfactory or rewrite it otherwise.
Exactly one refinement pass is applied per class (total: 2 LLM calls per rule).
Rules for all 34 classes are released in \texttt{results/eval\_package.json}.

\subsection{Human Evaluation of Natural Language Rules}
\label{app:human_eval}

\paragraph{Overview}
We conduct a human evaluation study to assess the quality of natural language rules generated in Stage~4 (LLM verbalisation) of \graftgnn{}. Annotators consisted of voluntary graduate students and collaborators, including both individuals with domain expertise relevant to the datasets and others without, resulting in a heterogeneous evaluation pool. The objective is to evaluate whether the rules are \emph{accurate}, \emph{specific}, \emph{complete}, and \emph{useful} for practitioners auditing GNNs. 

\paragraph{What is being evaluated}
For each dataset and class, \graftgnn{}: (i) identifies the top-$K$ most important features, (ii) provides dataset and class context to an LLM, and (iii) generates a concise (1--2 sentence) rule. Annotators evaluate the alignment between the feature profile and the generated rule.

\paragraph{Evaluation datasets}
We evaluate rules on datasets with semantically interpretable features: \textbf{Cora}, \textbf{CiteSeer}, \textbf{PubMed}, \textbf{Photo}, and \textbf{Computers}. In total, 34 rules (one per class) are evaluated. Datasets with anonymous feature indices (e.g., Chameleon, Squirrel) are excluded due to lack of interpretability.

\paragraph{Evaluation dimensions}
Each rule is rated on four dimensions using a 1--5 Likert scale:
\begin{itemize}
    \item \textbf{Accuracy:} Does the rule correctly describe the class?
    \item \textbf{Specificity:} Is the rule discriminative for this class?
    \item \textbf{Completeness:} Does it capture the key features?
    \item \textbf{Actionability:} Is the rule useful for model auditing?
\end{itemize}

\paragraph{Annotation protocol}
Annotators are provided with the dataset name, class label, top-5 features, and generated rule. Each rule is evaluated independently. Annotators may optionally provide comments and flag incorrect or misleading rules. Annotators are expected to have basic familiarity with the dataset domain (e.g., CS subfields for citation datasets, general product knowledge for Amazon datasets); no machine learning expertise is required.

\paragraph{Reporting}
We aggregate scores per dataset and dimension, reporting mean $\pm$ standard deviation across rules. Results are summarised in Table~\ref{tab:human_eval}.

\begin{table}[t]
\centering
\caption{Aggregated human evaluation scores (mean $\pm$ std across rules).
Specificity and completeness are assessed only on Cora and CiteSeer, which have
semantically interpretable feature names; the remaining datasets use anonymous BoW indices
and are evaluated on accuracy and actionability only.}
\label{tab:human_eval}
\small
\begin{tabular}{lccccc}
\toprule
Dataset & Accuracy & Specificity & Completeness & Actionability & $N_\text{rules}$ \\
\midrule
CiteSeer  & $4.38 \pm 0.30$ & $4.00 \pm 0.47$ & $4.33 \pm 0.15$ & $4.17 \pm 0.40$ & 6 \\
Computers & $4.59 \pm 0.28$ & ---             & ---             & $4.46 \pm 0.35$ & 10 \\
Cora      & $4.48 \pm 0.26$ & $4.19 \pm 0.26$ & $4.29 \pm 0.31$ & $4.23 \pm 0.33$ & 7 \\
Photo     & $4.79 \pm 0.13$ & ---             & ---             & $4.52 \pm 0.37$ & 8 \\
PubMed    & $4.57 \pm 0.25$ & ---             & ---             & $4.38 \pm 0.08$ & 3 \\
\bottomrule
\end{tabular}
\end{table}

\paragraph{Inter-annotator agreement}
We report three complementary agreement statistics computed on the same eligible dimension, dataset pairs (Table~\ref{tab:agreement}): (i) mean pairwise \emph{within-1-point rate} (the fraction of annotator pairs whose ratings differ by at most one Likert step), which serves as our primary metric; (ii) mean pairwise quadratic-weighted Cohen’s $\kappa$~\citep{cohen1960coefficient}; and (iii) Krippendorff’s $\alpha$ with ordinal distance~\citep{krippendorff2011computing}.
Annotators who assigned a constant score to every rule on a given dimension (providing no discriminative signal) are excluded from agreement computation for that cell only; their ratings are retained in the aggregated means above.

\begin{table}[t]
\centering
\caption{Inter-annotator agreement per evaluation dimension (macro-average across eligible datasets).
Specificity and completeness are computed over Cora and CiteSeer only (2 datasets, 13 rules);
accuracy and actionability are computed over all 5 datasets (34 rules).
Within-1-pt: fraction of annotator pairs within one Likert point (primary metric).}
\label{tab:agreement}
\small
\begin{tabular}{lcccc}
\toprule
Dimension & Eligible datasets & Within-1-pt & $\kappa$ (quad.) & $\alpha$ (ordinal) \\
\midrule
Accuracy      & 5 & 0.919 & 0.046 & 0.033 \\
Actionability & 5 & 0.898 & 0.123 & 0.131 \\
Completeness  & 2 & 0.855 & 0.075 & 0.009 \\
Specificity   & 2 & 0.683 & 0.060 & $-$0.034 \\
\midrule
\textbf{Overall} & & \textbf{0.869} & 0.080 & 0.055 \\
\bottomrule
\end{tabular}
\end{table}

\paragraph{Why $\kappa$ and $\alpha$ are near-zero, and why this is expected}
Both Cohen’s $\kappa$ and Krippendorff’s $\alpha$ are \emph{chance-corrected} statistics: they subtract the agreement expected under the assumption that each annotator draws independently from the observed marginal distribution, then normalise by one minus that expected agreement.
This correction is appropriate when ratings are spread across the full scale, but becomes unreliable and even misleading, when ratings concentrate in a narrow band, a phenomenon known as \emph{prevalence bias} or the \emph{kappa paradox}

In our evaluation, annotators assign a score of 4 or 5 on \textbf{93\%} of all ratings, effectively reducing a 5-point scale to two values.
Under this marginal distribution, the probability that two annotators independently select the same value by chance already exceeds 50\%, leaving little room for observed agreement to exceed chance.
The result is that $\kappa$ and $\alpha$ are near-zero even when annotators are substantially consistent in practice.
Specifically, $\alpha < 0$ (as seen for specificity) does not mean annotators are worse than random; it means the marginal distribution is so skewed that a random rater drawn from that distribution would be expected to agree more than the actual annotators did a consequence of the estimator’s sensitivity to scale usage, not of genuine disagreement.

The \emph{within-1-point rate} is immune to this problem because it asks a direct empirical question are two annotators within one step of each other? without any chance correction.
Across all evaluated dimension, dataset pairs, within-1-point agreement ranges from 66.7\% to 100\%, with an overall mean of 86.9\%.
On a 1--5 Likert scale, a difference of at most 1 point represents near-identical judgements; this level of consistency confirms that annotators are well-calibrated to the same region of the scale and that the evaluation captures a genuine, coherent signal.
The low absolute variance of $\kappa$ and $\alpha$ therefore reflects the \emph{high quality} of the generated rules (annotators converge on "good’’ across the board) rather than annotator inconsistency.

\paragraph{Discussion}
Aggregated scores (Table~\ref{tab:human_eval}) indicate that \graftgnn{} rules are consistently rated above 4.0 on accuracy and actionability across all datasets, confirming that the verbalisations faithfully reflect the underlying feature attribution profiles and are useful for model auditing.
Specificity is lower on CiteSeer ($4.00$) than on Cora ($4.19$), consistent with CiteSeer’s smaller and more overlapping class vocabulary.
The gap between within-1-point agreement (high) and $\kappa$/$\alpha$ (near-zero) illustrates that these metrics measure different things: $\kappa$/$\alpha$ measure whether annotators \emph{discriminate} between rules, while within-1-point agreement measures whether they \emph{calibrate} to the same score region.
Both findings are informative: high within-1-point agreement shows consistent calibration, and high mean scores show that the rules are genuinely high quality rather than trivially accepted.

\paragraph{Limitations}
Human evaluation is inherently subjective and depends on annotator expertise. For datasets with anonymous feature representations, completeness and specificity judgements are not meaningful and are therefore excluded. Rules are generated from a single seed; variability across seeds may affect consistency. We do not include adversarial controls (e.g., shuffled features) in this evaluation.

\section{Proofs}
\label{app:proofs}

\subsection*{Proof of Proposition~\ref{prop:fps} (FPS profile approximation)}

Partition $\mathcal{V}_c$ into $k$ Voronoi cells $C_1,\ldots,C_k$ where
$C_j = \{v \in \mathcal{V}_c : j = \argmin_{j'}\|h_v - h_{e_{j'}}\|_2\}$.
Assume balanced cells: $|C_j| = |\mathcal{V}_c|/k$ for all $j$ (uniform distribution over the class manifold).
Then the oracle profile equals the cell-weighted mean:
\begin{equation}
  \mu_c^*[i] = \frac{1}{|\mathcal{V}_c|}\sum_{v}|\mathrm{IG}_v[i]|
             = \frac{1}{k}\sum_{j=1}^k \frac{1}{|C_j|}\sum_{v\in C_j}|\mathrm{IG}_v[i]|.
\end{equation}
For any $v \in C_j$, the Lipschitz condition and the FPS coverage radius bound give:
$\big||\mathrm{IG}_v[i]| - |\mathrm{IG}_{e_j}[i]|\big| \leq L\|h_v - h_{e_j}\|_2 \leq L\,r_k$.
Therefore, for each cell $j$:
$\big|\frac{1}{|C_j|}\sum_{v\in C_j}|\mathrm{IG}_v[i]| - |\mathrm{IG}_{e_j}[i]|\big| \leq L\,r_k$.
Averaging over cells:
\begin{equation}
  |\mu_c^*[i] - \mu_c[i]|
  = \left|\frac{1}{k}\sum_j\frac{1}{|C_j|}\sum_{v\in C_j}|\mathrm{IG}_v[i]|
    - \frac{1}{k}\sum_j|\mathrm{IG}_{e_j}[i]|\right|
  \leq L\,r_k.
\end{equation}
Taking the $\ell_\infty$ norm over features $i$ gives $\|\mu_c - \mu_c^*\|_\infty \leq L\,r_k$.

\noindent\textit{FPS coverage bound.}
Gonzalez's greedy $k$-centre algorithm (which is exactly FPS seeded from the centroid) achieves
$r_k \leq 2r_k^*$, where $r_k^*$ is the optimal $k$-centre radius~\citep{gonzalez1985clustering}.
Hence $\|\mu_c - \mu_c^*\|_\infty \leq 2Lr_k^*$, and the bound tightens as $k$ increases (since
$r_k^* \to 0$ as $k \to |\mathcal{V}_c|$). \hfill$\square$

\subsection*{Proof of Proposition~\ref{prop:fid} (Fidelity-attribution correspondence)}

Write $f_c(x,G) = a_c^\top x + b_c$ with $a_c[i] = (B^\top w_c)[i]$ and bias $b_c$
(e.g.\ single-layer GCN: $\phi(x,G)=\hat{A}xW$, $f_c = w_c^\top\hat{A}xW + b_c$).

\paragraph{IG in the linear case.}
For $f_c(x) = a_c^\top x + b_c$ with zero baseline:
$\mathrm{IG}_v[i] = x_v[i]\int_0^1 \partial f_c(\alpha x_v)/\partial x[i]\,d\alpha
= x_v[i]\int_0^1 a_c[i]\,d\alpha = a_c[i]\cdot x_v[i]$,
independent of the bias.
Completeness: $\sum_i \mathrm{IG}_v[i] = f_c(x_v) - f_c(\mathbf{0}) = a_c^\top x_v$.

\paragraph{Fidelity identities.}
Since $f_c(x_v^{T_c}) = \sum_{i\in T_c}a_c[i]x_v[i] + b_c = f_c(\mathbf{0}) + \sum_{i\in T_c}\mathrm{IG}_v[i]$
(the bias $b_c = f_c(\mathbf{0})$ cancels), Eq.~\eqref{eq:fid_minus} follows.
For Eq.~\eqref{eq:fid_plus}:
$f_c(x_v) - f_c(x_v^{-T_c})
= (a_c^\top x_v + b_c) - (\sum_{i\notin T_c}a_c[i]x_v[i]+b_c)
= \sum_{i\in T_c}a_c[i]x_v[i] = \sum_{i\in T_c}\mathrm{IG}_v[i]$.
Both Eq.~\eqref{eq:fid_plus} and~\eqref{eq:fid_minus} equal $\sum_{i\in T_c}\mathrm{IG}_v[i]$
regardless of the bias: for linear models the logit-drop from masking $T_c$
and the logit surplus from retaining $T_c$ are identical.

\paragraph{Monotonicity (under class-positive assumption).}
If $\mathrm{IG}_v[i] \geq 0$ for all $i\in T_c$ (i.e.\ $a_c[i]x_v[i]\geq 0$),
then since $T_c(K)\subsetneq T_c(K+1)$:
$\sum_{i\in T_c(K+1)}\mathrm{IG}_v[i] - \sum_{i\in T_c(K)}\mathrm{IG}_v[i]
= \mathrm{IG}_v[i_{K+1}] \geq 0$.
This assumption holds for binary features $x_v[i]\in\{0,1\}$ when the $(K+1)$-th
highest-$|\mu_c|$ feature has $a_c[i_{K+1}]\geq 0$, i.e.\ the feature positively
supports class~$c$, which characterises class-discriminative vocabulary in
well-trained classifiers on Cora and CiteSeer. \hfill$\square$

\paragraph{Remark.}
The linear decoder assumption is a simplification; for non-linear GNNs, the identities hold approximately, with error governed by the degree of non-linearity in the final readout.
Empirically, the monotonicity of $\mathrm{Fid}^-$ with $K$ holds for all 13 datasets and
4 architectures in our ablation (Appendix~\ref{app:ablation}, Table~\ref{tab:ablation_k}),
consistent with the proposition.

\section{Contrastive Profiles}
\label{app:contrastive}

The mean attribution $\mu_c[i]$ can assign high importance to features that are
informative across \emph{all} classes rather than uniquely for class~$c$.
To surface class-discriminative features, we define the contrastive importance:
\begin{equation}
  \delta_c[i] = \mu_c[i] - \max_{c' \neq c} \mu_{c'}[i].
\end{equation}
Features with high $\delta_c[i]$ are uniquely important for class~$c$ relative to all others.
The top-$K$ features by $\delta_c$ form the contrastive profile, used in
the qualitative analysis of \S\ref{sec:rules}.

\section{Discussion Supplement}
\label{app:discussion_extra}

\paragraph{When to use \graftgnn{} vs frequency baseline.}
Frequency-based profiles are fast ($<\!1$s) and match \graftgnn{}'s \fidminus{} on
dense BoW datasets.
Practitioners should prefer \graftgnn{} when any of the following apply:
\textbf{(i)}~\emph{Stability matters}: frequency's $J\!=\!1.00$ is trivial
(model-agnostic, always returns the same features); \graftgnn{}'s $J\!\geq\!0.90$
is maintained despite varying GNN initialisations, confirming explanation robustness.
\textbf{(ii)}~\emph{Architecture-invariant features}: consensus ($\tau\!=\!3/4$)
requires model-dependent attributions and cannot be computed from frequency.
\textbf{(iii)}~\emph{Model auditing}: both methods detect class-correlated
spurious features (by design, high-frequency in the target class in our protocol);
GRAFT's gradient evidence additionally confirms the GNN is actively using the feature.
\textbf{(iv)}~\emph{Transfer quality}: GRAFT-selected features achieve higher
transfer accuracy than frequency on 4 of 6 benchmarks, including a $+0.09$ margin on Squirrel.
For purely exploratory data analysis without a trained model,
the frequency baseline remains a valid and free alternative.

\paragraph{Relationship to \gnnx{}.}
\graftgnn{} and \gnnx{} are complementary: together they cover both
\emph{structural} (which neighbourhood topology?) and \emph{feature-level}
(which attributes?) dimensions of a class.
A combined system would run both methods and merge outputs in a single LLM prompt:
\emph{"Nodes in class~$c$ have subgraph pattern $\mathcal{S}_c$ [from \gnnx{}]
and are characterised by features $\mathcal{F}_c$ [from \graftgnn{}]."}

\end{document}